\documentclass{article}

 \usepackage[dblblindworkshop, final]{neurips_2026}
\workshoptitle{Evaluation of Interactive Agents}

\usepackage[utf8]{inputenc} 
\usepackage[T1]{fontenc}    
\usepackage{hyperref}       
\usepackage{url}            
\usepackage{booktabs}       
\usepackage{amsfonts}       
\usepackage{nicefrac}       
\usepackage{microtype}      
\usepackage{xcolor}         
\usepackage{graphicx}
\usepackage{amsmath}
\usepackage{amsthm}
\usepackage{algorithm}
\usepackage{algorithmic}
\usepackage{mathtools}
\usepackage[inline]{enumitem}
\usepackage{tabularx}
\usepackage{multirow}
\usepackage{algorithm}
\usepackage{algorithmic}
\usepackage{cuted}
\usepackage{wrapfig}
\usepackage{caption}
\usepackage{tabularx}
\usepackage{multirow}
\usepackage[table]{xcolor}
\usepackage{tcolorbox}
\usepackage{graphicx}
\usepackage{float} 
\tcbuselibrary{listings}
\tcbuselibrary{skins, breakable}

\tcbset{
  promptbox/.style={
    colback=gray!10, colframe=gray!50,
    fonttitle=\bfseries, left=6pt, right=6pt, top=6pt, bottom=6pt
  },
  insertbox/.style={
    colback=#1!10, colframe=#1!60,
    fonttitle=\bfseries\small, left=4pt, right=4pt, top=4pt, bottom=4pt,
    boxrule=1pt
  }
}
\usepackage{wrapfig}
\usepackage{needspace}
\usepackage{booktabs}
\usepackage{makecell}

\newcommand{\hosein}[1]{\textcolor{green}{[Hosein: #1]}}

\newcommand{\ignore}[1]{{}}
\newcommand{\todo}[1]{\textcolor{red}{ToDo: #1}}
\definecolor{datasetrow}{RGB}{220, 230, 245} 

\usepackage{makecell}

\title{Ask Precisely, Assume Wisely:\\ Interactive Evaluation of Ambiguity Resolution in Large Language Models}

\author{%
\textbf{Avik Dutta} \quad
\textbf{Aayush Kumar} \quad
\textbf{Zahra Ashktorab} \\[6pt]
\textbf{Hosein Hasanbeig} \quad
\textbf{Austin Z. Henley} \quad
\textbf{Gustavo Soares} \\[10pt]
Microsoft Corporation
}

\begin{document}

\maketitle

\begin{abstract}
Ambiguity is a core challenge in human--AI interaction, where models must decide when to make assumptions and when to seek clarification. Existing approaches to ambiguity evaluation largely focus on coverage over multiple interpretations using static metrics, offering limited insight into how efficiently a model converges toward a single user-intended interpretation through interaction. We introduce a preference-grounded simulated user framework for controlled, multi-turn evaluation of ambiguity resolution, grounding simulator responses strictly in user disambiguation preferences to avoid hallucination and over-disclosure. Alongside this, we define three observable dissatisfaction signals --- incorrect assumptions, repeating clarifications, and irrelevant clarifications --- that characterize interaction quality beyond task accuracy. Our analysis reveals a clear trade-off: models that frequently seek clarification tend to over-clarify and prolong interactions, while assumption-driven models are more prone to misaligning with user intent. We demonstrate that system prompts optimized on these dissatisfaction signals achieve a Pareto-optimal balance across all three metrics, reducing overall error rates by up to 7.65\%, underscoring the value of our framework in designing agents that ask precisely and assume wisely.
\end{abstract}

\ignore{Much of the friction in human–agent interaction arises not from computational errors but from misaligned assumptions, unnecessary or repeating clarifications. To meaningfully study these failures, we need controlled ambiguity, realistic user behavior, and a way to quantify the user's dissatisfaction when an agent’s reasoning or questions moves off course. This work is motivated by the need to evaluate and improve how agents handle such ambiguity. By introducing a structured simulated user which helps navigate the models (1) ambiguous tasks, (2) a structured user proxy simulating diverse user profiles and backgrounds \hosein{to converse with the agent}, and (3) a dissatisfaction metric, we aim to surface the conversational behaviors that distinguish helpful agents from ones that inadvertently frustrate users.
    \hosein{can replace the previous sentence: To identify conversational behaviors that distinguish helpful agents from those that inadvertently frustrate users, we introduce three key elements: (1) a structured set of ambiguous tasks, (2) a user proxy simulating diverse user profiles and backgrounds that agents can converse with, and (3) a dissatisfaction metric.}
    \hosein{We should also mention the main findings here.}}

\section{Introduction}

\begin{figure*}[t]
  \includegraphics[width=\textwidth]{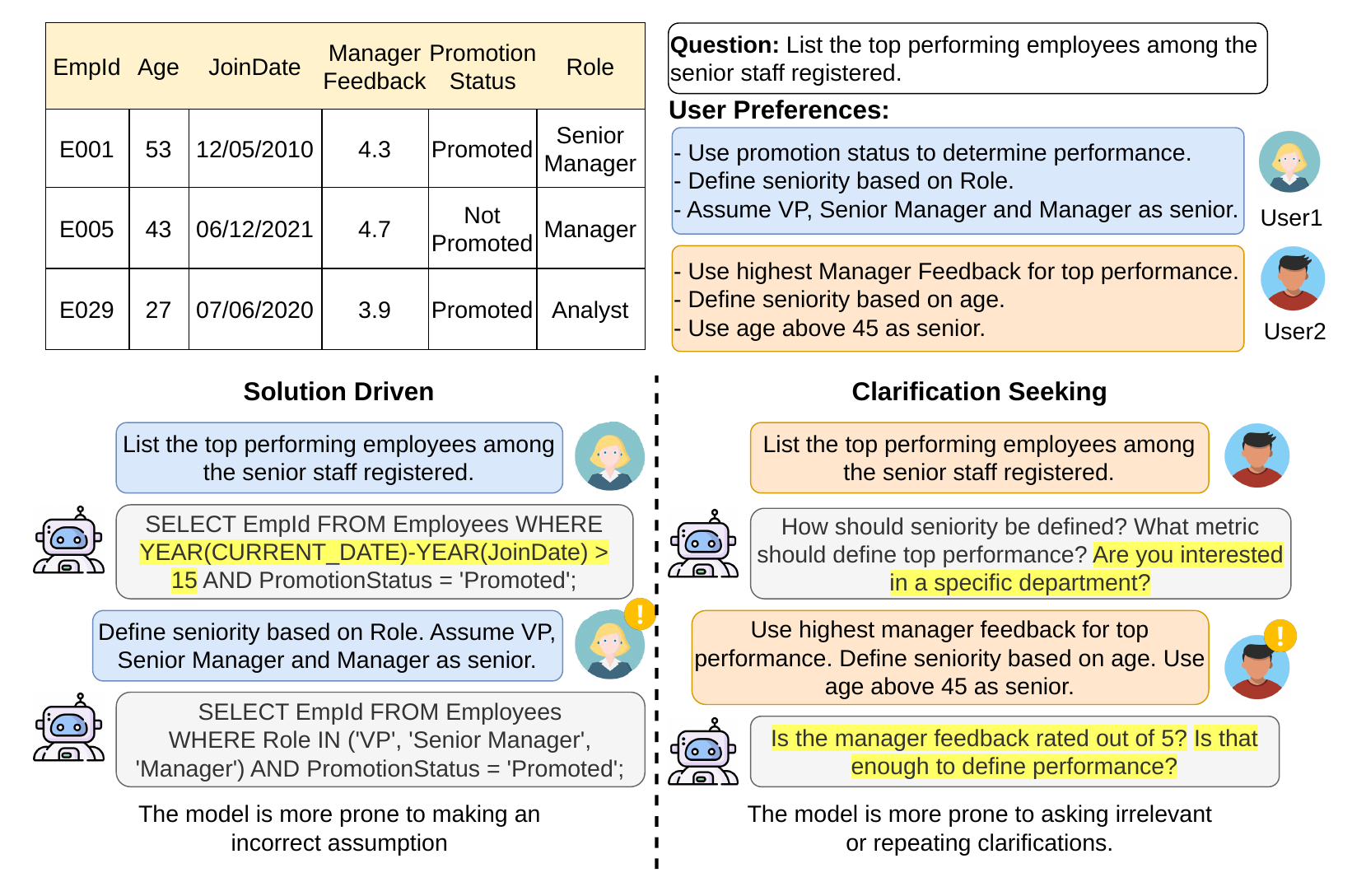}
  \caption{
  \textbf{Demonstrates different ambiguity resolution styles and signals of user dissatisfaction during interaction.}
  Users may choose different resolution strategies towards a given ambiguity.
  Solution-driven models often make incorrect assumptions while clarification-seeking models ask repeating or irrelevant clarifications.
  }
  \label{fig:motivation_diagram}
\end{figure*}

\ignore{
Underspecified user instructions are quite common and a natural tendency in real-world human-AI conversations \cite{herlihy2024overcomingmiscalibratedconversationalpriors}, termed as "the principle of least effort" \cite{zipf1949least}.
In such cases, engaging in interaction to better understand user intent becomes essential. With recent advances in large language models (LLMs), interactive AI assistants (e.g., ChatGPT, Claude, LLaMA-2) have become increasingly prevalent in everyday use. While these models are widely recognized for their ability to solve complex tasks, recent efforts have also focused on improving their collaborative capabilities. However, increased collaborativeness does not necessarily translate to effective ambiguity resolution, as it does not guarantee that models will appropriately seek clarifications when needed.

We observed that only in approximately 30\% of the times in Ambrosia, has the model been able to correctly align, or rather guess the correct solution. This demonstrates that 70\% of users remain unsatisfied with the LLM's solution.

Our contributions can be listed as follows:
\begin{itemize}
    \item A simulated user model that enables controlled multi-turn interactions between the user and the agent. The model is grounded in a set of preference specifications and instantiated via different grounding recipes, each representing a distinct user profile and intent for resolving ambiguity. This design allows us to systematically vary user behavior and control how conversations unfold, enabling scalable and reproducible evaluation of agent performance under diverse user preferences.
    \item This simulated user design allows us to derive key observable indicators of user dissatisfaction, such as incorrect assumptions, redundant clarifications, and irrelevant queries, motivated by prior work and empirical analysis.
    \item These signals captured over a conversation reveals a clear trade-off in models, where clarification seeking models are prone to asking un-grounded or repeating clarification, thereby prolonging conversation. On the other hand, assumption driven models are more prone to having a different interpretation to the query than the user.
\end{itemize}

\todo{Points/Experiments that should be added:
\begin{itemize}
    \item Support the design of Interactive Eval better: explain how it is not biased towards a specific failure. Justify the metric design for Irrelevant clarification. Explain that this metric is well rounded on what it measures. Justify the applicability of the assumptions under which this metric operates. Explain why each failure component can be added under the same scale.
    \item Need to justify metric diligence by correlating with user studies. This paper currently lacks this justification.
    \item Add an experiment on User-Proxy design, demonstrating that it is reliable and replicable.
    \item See if a model can be finetuned to show benefits of this metric.
    \item show statistical significance test on one of the run.
\end{itemize}
}
}

Ambiguity is a fundamental challenge in human--AI interaction. Real-world user queries are often under-specified, requiring language models (LMs) to infer missing intent or engage in clarification through interaction \cite{keyvan2022approach}. 
While recent LLMs have demonstrated strong capabilities across reasoning and conversational tasks, effectively resolving ambiguity remains difficult. 
Models that aggressively assume a single interpretation risk misunderstanding user intent \cite{berant2025learning}, whereas models that over-rely on clarification can prolong interactions through repetitive or irrelevant questions \cite{acikgoz2026mac,andukuri2024star}. 
Designing agents that can \emph{ask precisely and assume wisely} therefore remains an open problem in interactive language systems.

Despite growing interest in interactive evaluation, existing approaches remain limited in two important ways. First, evaluation frameworks largely measure final task success, overlooking the quality and efficiency of the conversational process used to arrive at the answer \cite{zhang2023clarifynecessaryresolvingambiguity, min2020ambigqaansweringambiguousopendomain}. Second, many interactive evaluations rely on LLM-only user simulators to facilitate multi-turn interaction. While flexible, these proxies are difficult to control and reproduce, often prone to hallucination and over-disclosure of information \cite{dou-etal-2025-simulatorarena, seshadri2026lostsimulationllmsimulatedusers}.


We address both limitations by introducing a \textbf{preference-grounded user simulator} for evaluating ambiguity resolution across multiple turns.
We formalize each ambiguous query through a user disambiguation preference set as the minimal set of specifications required to uniquely identify the user's intended interpretation. 
This simulator restricts its responses strictly to these preferences, enabling reproducible and controlled evaluation without risk of information leakage. Derived from this are three observable \textbf{dissatisfaction signals} — incorrect assumptions, repeating clarifications, and irrelevant clarifications — that together provide a more nuanced view of how efficiently a model navigates ambiguity during interaction.

Our analysis spanning ambiguous question answering, semantic parsing and text-to-visualization benchmarks reveal a clear trade-off: assumption-driven strategies have higher frequency of wrong attempts while clarification-seeking models prolong conversations while asking ill-scoped questions.
By prompt-optimizing on our dissatisfaction signals (DisPO), we achieve pareto optimal balance, leading to cleaner interactions without sacrificing task success.
Our overall contributions include:
\begin{itemize}[itemsep=0pt, topsep=2pt, parsep=0pt, partopsep=0pt]
    \item We propose a preference-grounded user simulator for controlled and reproducible evaluation towards ambiguity resolution in a multi-turn setting.
    \item We introduce three dissatisfaction signals that correlate to inefficient ambiguity handling in conversations.
    \item We demonstrate a clear trade-off between assumption-driven and clarification-seeking strategies, where optimization on our framework yields systems that ask only necessary and precise questions.
\end{itemize}


\section{Problem Formulation}
We formalize ambiguity and define how different disambiguation preferences correspond to different user interpretations. We then introduce our simulated user framework, which grounds interactions on these preferences, and describe how it enables measuring user dissatisfaction signals as indicators of ambiguity resolution effectiveness.

\subsection{User Disambiguation Preferences}
User Disambiguation Preferences are resolution strategies that guide an ambiguous query toward a specific intended interpretation
A query \(q\) is \textit{ambiguous} if the cardinality of its solution space satisfies \(|\mathcal{A}(q)| > 1\), where \(\mathcal{A}(q)\) denotes the set of valid solutions consistent with the query.
For example, in Figure~\ref{fig:motivation_diagram}, \textit{seniority} could be interpreted in multiple ways: age, organizational role, or joining date, leading to multiple solutions.
To resolve ambiguity toward a target solution \(a \in \mathcal{A}(q)\), we introduce a user disambiguation preference set \(P = \{p_1, p_2, \dots, p_n\}\),such that \(q \circ P\) admits a unique solution \(a\). 
We require \(P\) to satisfy: (i) \textit{completeness}, where removing any \(p_i \in P\) reintroduces ambiguity; (ii) \textit{non-redundancy}, where no specification \(p_i\) can be removed or changed without altering solution \(a\); and (iii) \textit{disentangled scope}, where each \(p_i \in P\) resolves a distinct aspect of ambiguity, either independently or conditioned on previously resolved ambiguities.
For example, in Figure~\ref{fig:motivation_diagram}, the user preferences resolve: (1) how \textit{performance} should be evaluated; (2) what constitutes \textit{seniority}; (3) the threshold or parameter associated with that definition (e.g., age threshold or role hierarchy). 
Each of these are disjoint decision points when a solution is computed.
Different assignments of \(P\) therefore model different user interpretation toward a specific task. 
The cardinality \(|P|\) symbolizes the extent of uncertainty characterized by the number of decision points at which the model encounters multiple plausible interpretations.
This definition extends to queries whose interpretation space is unbounded. In such cases, $P$ contains only the specifications the user cares about, and every aspect left out is one for which any reasonable assumption is acceptable. $P$ is therefore finite for a given user, even when the space of possible interpretations is not.

\subsection{Preference-Grounded User Simulator}

\begin{wrapfigure}{r}{0.38\columnwidth}
    \centering
    \includegraphics[width=\linewidth]{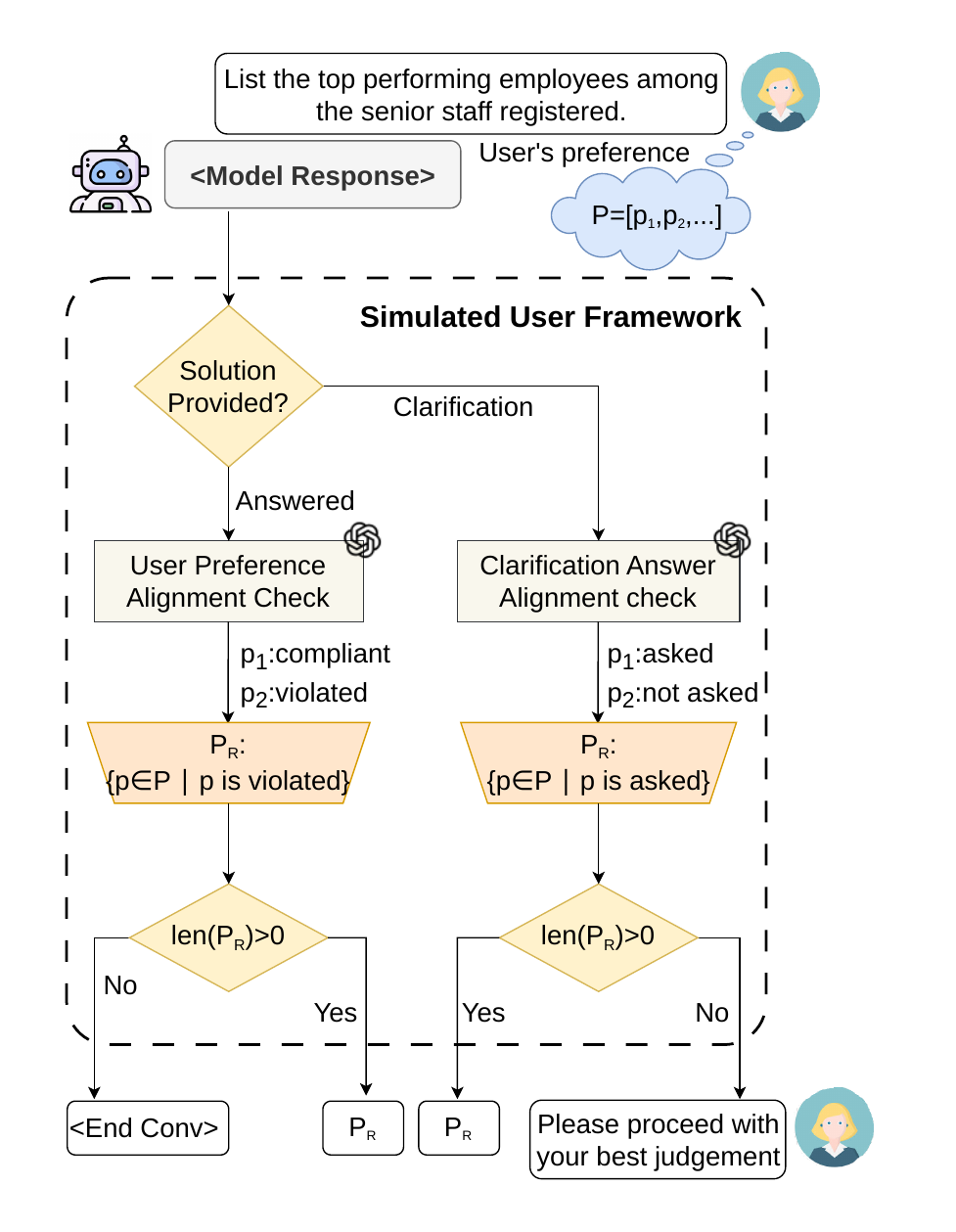}
    \caption{
        Preference-Grounded User Simulator that \textit{rectifies} or \textit{clarifies} based on preference set $P$.
    }
    \label{fig:simulated_user_design}
\end{wrapfigure}

We propose a user simulator that generates responses grounded on the disambiguation preference set \(P\) corresponding a given interpretation.
Prior work on multi-turn evaluation commonly relies on a single LLM call to serve as user simulator.  
While they can produce natural conversations \cite{dou-etal-2025-simulatorarena}, the interactions remain difficult to control and reproduce consistently across evaluations, and are prone to hallucinated user behavior and unnecessary information disclosure beyond what is required to resolve the ambiguity.

To address these limitations, we design a simulator where the user vocabulary is restricted to the preference set \(P\) (see Figure~\ref{fig:simulated_user_design}).
At each interaction turn $t$, the evaluated agent may either provide a candidate solution or seek clarification. If a solution is proposed, the framework checks whether the generated answer satisfies all specifications in the user preference set $P$. Any unresolved or incorrectly assumed ambiguity is rectified by responding with: $\{p\in P | p:violated\}$. 
If the agent instead asks a clarification question, the framework determines whether the question meaningfully targets a specific ambiguity dimension resolved by some \(p_i\). In such cases, the simulated user returns the associated clarification response $\{p\in P | p:asked\}$; otherwise, the user responds with ``\textit{Please proceed with your best judgment}'' \cite{vijayvargiya2025interactiveagentsovercomeambiguity}, if the set is empty. 
The conversation terminates when agent's solution aligns with $P$.
Our goal is not to faithfully emulate real users, but to enable controlled information injection into interaction trajectories for systematically analyzing ambiguity handling by models \cite{vijayvargiya2025interactiveagentsovercomeambiguity}.

\subsection{Measuring Ambiguity Resolution}
\label{sec:dissat_formulas}
Beyond task success, our framework measures how efficiently the model resolves ambiguity toward the user's intended interpretation, focusing on three dissatisfaction signals: incorrect assumptions, repeated clarifications, and irrelevant clarifications \cite{inproceedings, shea2026sagetopdownbottomupknowledgegrounded}. 
We formalize these dissatisfaction signals through two indicator functions.
The \textit{Answer} indicator function \(A(t,p)=\mathbb{I}\!\left[p \in P \land (p=\texttt{violated} \lor p=\texttt{asked})\right]\), takes the value \(1\) if the user reveals preference \(p \in P\) at turn \(t\) and \(0\) otherwise.
Similarly, we define a \textit{Clarification} indicator function \(Q(t,p)=\mathbb{I}\!\left[ p \in \{P \cup \bar{P}\} \land p=\texttt{asked}\right]\), which takes the value \(1\) if the agent asks clarification at turn \(t\) for any preference $p$ that may or may not be part of $P$, else \(0\).
$\overline{P}$ denotes complement of $P$, suggesting those questions that resolve a preference $p\notin P$, i.e. irrelevant towards resolving ambiguity.

\paragraph{Incorrect Assumption.} 
We count assumptions that do not match the user's preference throughout the conversation. This is tracked by counting the number of repair utterances from the user\cite{walker1997paradiseframeworkevaluatingspoken} when clarification is not asked.
$$
\mathrm{IncAssmp}
=
\sum_{t=1}^{\tau}
\sum_{p \in P}
A(t,p)\,
\mathbf{1}\!\left\{ Q(t,p)=0 \right\}
$$

\paragraph{Repeating Clarification.}
We capture the number of times the agent asks a clarification question, whether relevant to \(P\) or not, that has already been previously asked or resolved by the user in its response earlier, thereby introducing unnecessary conversational repetition and user frustration.
$$
\begin{aligned}
\mathrm{RepClrf} =
& \sum_{t=1}^{\tau}\sum_{p \in P \cup \overline{P}}
Q(t,p)\,
\mathbf{1}\Biggl\{ 
& \sum_{s=1}^{t-1}
\left(Q(s,p)+A(s,p)\right) > 0
\Biggr\}
\end{aligned}
$$

\paragraph{Irrelevant Clarification.}
We measure how often the agent asks clarification questions unrelated to resolving the intended ambiguity.
Formally, this corresponds to questions \(Q(t,p)\) where \(p \in \overline{P}\).

$$
\mathrm{IrrelClrf}= \sum_{t=1}^{\tau} \frac{\sum_{p \in \overline{P}} Q (t,p)}{\mathcal{I}(q)^{1-\frac{r(t)}{|P|}}}
$$

We scale each irrelevant question detected by $\mathcal{I}(q)^{1-\frac{r(t)}{|P|}}$, where $\mathcal{I}(q)$ denotes the number of sensible interpretations (cardinality of assignment space of $P$) to ambiguous query $q$, and $r(t)$ is a monotonic function that stores the count of preferences that were either revealed by the user or the agent got them right at turn $t$:
$r(t) \coloneqq \left| P_{\text{revealed}}(t) \right|, \quad P_{\text{revealed}}(t) \subseteq P $.
The scaling factor is added to allow agents room to ask clarification questions more freely earlier when the user intent is unclear, and be more restrictive when a certain amount of information has been revealed over the turns. During the course of the conversation as the simulated user reveals its preference, the solution space reduces by a multiplicative factor and so does the room for asking irrelevant clarifications.
Consequently, the penalty increases exponentially over the course of the interaction.

These metrics are not directly comparable, as they arise from different decision contexts and opportunity spaces. Incorrect assumptions occur when the agent chooses to act, while clarification-based errors occur when the agent chooses to ask. Repeating clarification depends on interaction history, whereas irrelevant clarification depends on the remaining ambiguity space. This creates a turn-wise asymmetry: early interactions allow many irrelevant clarifications but few repetitions, while later turns increase repetition likelihood and space for irrelevant queries shrink. To account for this, we normalize irrelevant clarification with lower penalties in early turns and higher penalties in later turns, aligning it with repeated clarification. For example, in a query with two interpretations and a single ambiguity $|P|=1$, two early irrelevant clarifications is equal to one repeated clarification.

\ignore{
\paragraph{DisSat Score.} By adding all the above metrics, we obtain what we call DisSat score, which reflects the overall dissatisfaction expressed by the user during a conversation with the agent attempting to solve $q$ on $d$.
$$
\mathrm{DisSat} = \sum_{t=1}^{\tau} \left[ \frac{1}{|P^c|}\frac{\sum_{p \notin P} Q_a (t,p)}{\kappa^{|P|-r(t)}} + \frac{1}{|P|}\sum_{p \in P} ReLU( \lvert A_u (t, p)\rvert - \lVert Q_a (t,p) \rVert ) \right]
$$

\todo{Drop the $|P^c|$ normalization and instead divide each factor by the opportunity that each could occur (gives a true 0..1). The current way would penalize long conversations? I have ideas, will put it into latex later.}

\todo{Claude suggested presenting it this way:}

\[
\widehat{\mathrm{Inc}}(\tau)=
\frac{
\sum_{t=1}^{\tau}\sum_{p \in P} A_u(t,p)\mathbf{1}\!\left\{Q_a(t,p)=0\right\}
}{
\sum_{t=1}^{\tau}\sum_{p\in P} A_u(t,p)+\epsilon
}
\]

\[
\widehat{\mathrm{Rep}}(\tau)=
\frac{
\sum_{t=1}^{\tau}\sum_{p \in P}
A_u(t,p)\mathbf{1}\!\left\{\sum_{s=1}^{t}Q_a(s,p)>1 \land Q_a(t,p)=1\right\}
}{
\sum_{t=1}^{\tau}\sum_{p\in P}Q_a(t,p)+\epsilon
}
\]

\[
\widehat{\mathrm{Irr}}(\tau)=
\frac{
\sum_{t=1}^{\tau}
\frac{\sum_{p\notin P}Q_a(t,p)}{\kappa^{|P|-r(t)}}
}{
\sum_{t=1}^{\tau}
\frac{\sum_{p\in P}Q_a(t,p)+\sum_{p\notin P}Q_a(t,p)}{\kappa^{|P|-r(t)}}
+\epsilon
}
\]

\[
\mathrm{DisSat}(\tau)=
\frac{1}{3}\left(
\widehat{\mathrm{Inc}}(\tau)+
\widehat{\mathrm{Rep}}(\tau)+
\widehat{\mathrm{Irr}}(\tau)
\right)
\]

The DisSat formulation is designed to be failure-type complete under the ambiguity-resolution setting. Each agent action at a turn must correspond to either (i) making an assumption, (ii) requesting clarification, or (iii) providing a compliant resolution. The three components (IncAssmp, RepClar, IrrelClar) partition these failure behaviors without overlap, preventing bias toward a single interaction style. \todo{Add citations from prior works to support metric completeness and coverage}.
}

\section{Dataset and Metrics}

\paragraph{AmbigQA} covers ambiguity in open-domain question answering tasks \cite{min2020ambigqaansweringambiguousopendomain}. Built on top of NaturalQuestions \cite{kwiatkowski-etal-2019-natural}, the dataset provides annotated disambiguation variants for each ambiguous query, which we leverage to construct user disambiguation preference sets. We split 1,142 unique queries into 3,649 variants, with an average of 3.2 interpretations per query and an $|P|_{avg}=$ 1.26. Given the open-ended nature of responses, we employ GPT-5 (reasoning) as LLM-judge \cite{zheng2023judgingllmasajudgemtbenchchatbot} to evaluate, instead of relying on lexical similarity metrics such as EDIT-F1 or BLEU \cite{bleu}, which are overly strict and have higher false positives.

\paragraph{Ambrosia} covers ambiguity in text-to-SQL semantic parsing tasks \cite{saparina2024ambrosiabenchmarkparsingambiguous}. The dataset captures diverse forms of ambiguity in tabular queries, including scope ambiguity, attachment ambiguity, and vagueness, along with their corresponding unambiguous variants, which are used to construct the preference set.
From 1,141 unique ambiguous queries, we build a total interaction set of 2,689 instances, with an average of 2.36 interpretations per query and $|P|_{avg}=$ 1.16. For evaluation, we measure execution accuracy by comparing the execution result of the predicted SQL query with that of the gold query.

\paragraph{nvBench2.0} covers ambiguous queries in \emph{Text2VIS} systems, where models generate visualizations from tabular data \cite{luo2026nvbench20resolvingambiguity}. 
It introduces ambiguity on existing table-query pairs from nvBench \cite{10.1145/3448016.3457261} and BIRD \cite{li2023llmservedatabaseinterface}, and provides stepwise disambiguation reasoning paths, which we leverage to construct user preference sets. From 747 unique ambiguous queries, we derive 2,251 interaction instances, with an average of 3.01 interpretations per query and $|P|_{avg}=$ 1.55. Following \cite{9617561, 10.1145/3448016.3457261}, we evaluate predictions by comparing the generated Vega-Lite specifications with the corresponding gold.

We expand each query into all possible interpretations, enabling ambiguity resolution to be evaluated over the complete interpretation space rather than being biased toward specific user preferences.
In addition to task success, we report the Clean Interaction Rate (CIR\%), which measures the percentage of benchmark queries whose conversations contain no \textit{Incorrect Assumption}, \textit{Repeating Clarification} and \textit{Irrelevant Clarification}.

\begin{table*}[t]
\footnotesize
\centering
\begin{tabularx}{\linewidth}{l *{6}{>{\centering\arraybackslash}X}}
\toprule
\textbf{Model} & \textbf{Acc(\%)$\uparrow$} & \textbf{|Conv|$\downarrow$} & \textbf{IncAssmp$\downarrow$} & \textbf{RepClrf$\downarrow$} & \textbf{IrrelClrf$\downarrow$} & \textbf{CIR(\%)$\uparrow$} \\
\midrule
\rowcolor{blue!10}
\multicolumn{7}{c}{\textbf{AmbigQA}} \\
\midrule

Phi-4 & 34.34 & 2.03 & 1.13 & 0.02 & 0.02 & 21.54 \\

DeepSeek-R1-Distill & 30.79 & 2.05 & 1.07 & 0.06 & 0.04 & 23.36 \\

Claude-Opus-4.6 & \textbf{65.11} & \textbf{1.85} & 0.88 & 0.05 & 0.02 & 30.50 \\

GPT-5.4 & 63.47 & 1.90 & 0.81 & 0.12 & 0.02 & \textbf{31.84} \\

\midrule
\rowcolor{blue!10}
\multicolumn{7}{c}{\textbf{Ambrosia}} \\
\midrule

Phi-4 & 44.55 & 1.92 & 0.98 & 0.01 & 0.00 & 34.81 \\

DeepSeek-R1-Distill & 45.15 & \textbf{1.74} & 0.79 & 0.00 & 0.00 & 36.26 \\

Claude-Opus-4.6 & 53.48 & 1.83 & 0.89 & 0.00 & 0.00 & \textbf{37.60} \\

GPT-5.4 & \textbf{66.20} & 1.84 & 0.89 & 0.03 & 0.00 & 36.04 \\

\midrule
\rowcolor{blue!10}
\multicolumn{7}{c}{\textbf{nvBench 2.0}} \\
\midrule

Phi-4 & 49.27 & 1.95 & 1.32 & 0.00 & 0.00 & 16.30 \\

DeepSeek-R1-Distill & 51.78 & 2.01 & 1.28 & 0.00 & 0.04 & 15.77 \\

Claude-Opus-4.6 & 69.44 & 1.84 & 1.17 & 0.00 & 0.00 & 19.24 \\


GPT-5.4 & \textbf{69.48} & \textbf{1.83} & 1.16 & 0.00 & 0.00 & \textbf{19.90} \\

\bottomrule
\end{tabularx}
\caption{
Reports task success, measured by accuracy Acc(\%), alongside interaction efficiency through Clean Interaction Rate CIR(\%) and average conversation length $|\text{Conv}|$, after the model aligns with the user’s intended interpretation of an ambiguous query.  
$\uparrow$ and $\downarrow$ denote metrics where higher and lower values are preferable, respectively. Closed-weight models generally achieve stronger overall performance and efficiency.
}
\label{tab:main_table_models}
\end{table*}


\section{Evaluation}

\begin{itemize}[itemsep=0pt, topsep=2pt, parsep=0pt, partopsep=0pt]
    \item[\bfseries RQ1] How different models and strategies effectively resolve ambiguity?
    \item[\bfseries RQ2] Can dissatisfaction signals reliably guide when agents should ask clarifying questions versus make assumptions?
    \item[\bfseries RQ3] How resolution strategies change with increasing ambiguity?
    \item[\bfseries RQ4] Can our user simulator overcome limitations of existing LLM-only baselines?
\end{itemize}

\subsection{Ambiguity Resolution in LLMs (RQ1)}
We evaluate LLMs and different clarification strategies by measuring their ability to resolve ambiguity through interactions with our user simulator.

\subsubsection{Evaluating Models}

\paragraph{Framework.}
We evaluate both open-weight: Phi-4 \cite{phi4_2024}, Deepseek-R1-Distill-Qwen-32B (abbrv. as Deepseek-R1-Distill) \cite{deepseekai2025deepseekr1incentivizingreasoningcapability}; and proprietary models: Claude-Opus-4.6 \cite{anthropic2026claudeopus46} and GPT-5.4 \cite{openai_gpt54_2026}. Each model interacts (upto 10 turns) with our user simulator. No explicit instruction is given to indicate that the query is ambiguous, and all models operate in a zero-shot setting (see Appendix~\ref{appx:prompts_used} for exact prompts).
We use GPT-5 (chat) throughout to drive our user simulator.

\paragraph{Results.}
In the absence of explicit indicators about query ambiguity, models exhibit an inherent solution-driven behavior and attempt to resolve queries in the first turn \cite{zhang2025modelingfutureconversationturns}. As a result, incorrect assumptions occur frequently, while repeating and irrelevant clarifications remain negligible (Table~\ref{tab:main_table_models}).
Models with shorter conversation lengths generally achieve higher CIR, suggesting a greater ability to infer the intended interpretation in the first turn. Despite user repair utterances following incorrect assumptions, open-weight models consistently underperform, particularly on the open-domain ambiguity benchmark AmbigQA, where their accuracy is nearly half that of closed-weight models, indicating weaker pre-training knowledge. NvBench2.0 is especially challenging for models to infer correctly, as reflected by its comparatively lower CIR, with models make at least one incorrect assumption in most interactions.



\subsubsection{Evaluating Resolution Strategies}


\begin{table*}[t]
\footnotesize
\centering
\begin{tabularx}{\linewidth}{l *{6}{>{\centering\arraybackslash}X}}
\toprule
\textbf{Strategy} & \textbf{Acc(\%)$\uparrow$} & \textbf{|Conv|$\downarrow$} & \textbf{IncAssmp$\downarrow$} & \textbf{RepClrf$\downarrow$} & \textbf{IrrelClrf$\downarrow$} & \textbf{CIR(\%)$\uparrow$} \\
\midrule
\rowcolor{blue!10}
\multicolumn{7}{c}{\textbf{AmbigQA}} \\
\midrule
None & 63.47 & 1.90 & 0.81 & 0.12 & 0.02 & 31.84 \\
Self-Ask & 65.72 & 1.96 & 0.85 & 0.14 & 0.02 & 30.25 \\
Intent-Sim & 62.02 & 1.85 & 0.86 & 0.07 & 0.02 & 29.60 \\
DisPO & 64.68 & 2.10 & 0.68 & 0.15 & 0.11 & \textbf{39.49} \\
Clrf & 59.80 & 4.20 & 0.28 & 1.83 & 0.91 & 24.61 \\
\midrule
\rowcolor{blue!10}
\multicolumn{7}{c}{\textbf{Ambrosia}} \\
\midrule
None & 66.20 & 1.84 & 0.89 & 0.03 & 0.00 & 36.04 \\
Self-Ask & 67.39 & 1.90 & 0.84 & 0.07 & 0.02 & 37.45 \\
Intent-Sim & 68.50 & 1.83 & 0.82 & 0.02 & 0.02 & 35.78 \\
DisPO & 68.72 & 2.00 & 0.68 & 0.09 & 0.10 & \textbf{44.78} \\
Clrf & 61.66 & 3.33 & 0.34 & 0.99 & 1.64 & 17.36 \\
\midrule
\rowcolor{blue!10}
\multicolumn{7}{c}{\textbf{nvBench 2.0}} \\
\midrule
None & 69.48 & 1.83 & 1.16 & 0.00 & 0.00 & 19.90 \\
Self-Ask & 69.35 & 1.95 & 1.06 & 0.04 & 0.11 & 26.03 \\
Intent-Sim & 68.50 & 1.91 & 1.13 & 0.03 & 0.04 & 20.61 \\
DisPO & 69.30 & 2.05 & 0.93 & 0.03 & 0.20 & \textbf{32.03} \\
Clrf & 70.06 & 2.30 & 0.73 & 0.08 & 0.60 & 29.63 \\
\bottomrule
\end{tabularx}
\caption{
Interaction efficiency for different ambiguity resolution strategies. 
\textit{DisPO} achieves a Pareto-optimal trade-off between incorrect assumptions and repeated or irrelevant clarifications.
}
\label{tab:main_table_strategy}
\end{table*}

\paragraph{Framework.}
In this setup, we pick one of the best performers i.e. GPT-5.4, and evaluate different ambiguity resolution strategies. For baselines, we use (i) \textbf{None}:  where no explicit guidance for handling ambiguity is provided; (ii) \textbf{Clrf}: model is instructed to ask clarification upon detecting ambiguity; (iii) \textbf{Self-Ask}: implements chain-of-thought to determine if and what clarification to ask --- "Is a follow-up question needed here?", adapted to shift focus from task decomposition \cite{press2023measuringnarrowingcompositionalitygap}; (iv) \textbf{Intent-Sim}: determines if clarification is necessary by estimating entropy over possible user interpretations \cite{zhang-choi-2025-clarify}. Clarification is asked when entropy exceeds 0.5 (sample size $S=8$, Temperature $T=1$).


\paragraph{Results.}
For a given model (GPT-5.4), accuracy remains largely stable across ambiguity resolution strategies after attaining user alignment, reflecting a capability ceiling that each model reaches regardless of how it resolves ambiguity (as empirically observed in Table~\ref{tab:main_table_strategy}).
What differs across strategies is the quality and efficiency of the interaction path used which is captured via CIR(\%).
Under \textsc{None}, models answer directly, resulting in shorter conversations but higher incorrect assumptions. In contrast, \textsc{Clrf} reduces incorrect assumptions by aggressively seeking clarification, but at the cost of longer interactions and increased repeated and irrelevant clarifications from nearly zero to average 0.97 and 1.05 respectively, revealing inefficiencies in unconstrained clarification-seeking. \textsc{Self-Ask} and \textsc{Intent-Sim} occupy a middle ground, maintaining shorter conversations while keeping repeated and irrelevant clarifications minimal, thereby achieving a more balanced trade-off between assumption risk and conversational overhead.

\subsection{Dissatisfaction-Guided Resolution (RQ2)}

We use \textbf{Dis}satisfaction-guided \textbf{P}rompt \textbf{O}ptimization (DisPO) to define instructions on when agents should ask clarifying questions and how precise those clarifications should be, using dissatisfaction signals collected from conversations \cite{cheng2024blackboxpromptoptimizationaligning, yang2023large}. The optimization leverages conversations generated by both extremes of solution-driven (\textit{None}) and clarification-seeking (\textit{Clrf}) models. We collected 100 random GPT-5.4 conversations from each benchmark (AmbigQA, Ambrosia, nvBench2.0) under both settings, yielding 600 conversations. Using GitHub Copilot CLI \cite{githubcopilot} in an automated review loop, we inspected each conversation alongside its dissatisfaction signal to identify recurring mistake patterns \cite{liu-etal-2025-user}, and iteratively refined the prompt to address them (exact prompt in Appendix~\ref{appx:prompts_used}). To reduce order sensitivity, we repeated the process with the conversations in reverse order and retained only the instructions common to both resulting prompts. We performed no further optimization on this sample to avoid overfitting, and the final DisPO prompt was fixed.
Table~\ref{tab:main_table_strategy} shows that DisPO achieves the highest CIR\% across all benchmarks, establishing a target for models to strive toward in achieving efficient ambiguity resolution.

\subsection{Impact of Increasing Ambiguity (RQ3)}

\begin{figure*}[t]
  \includegraphics[width=\textwidth]{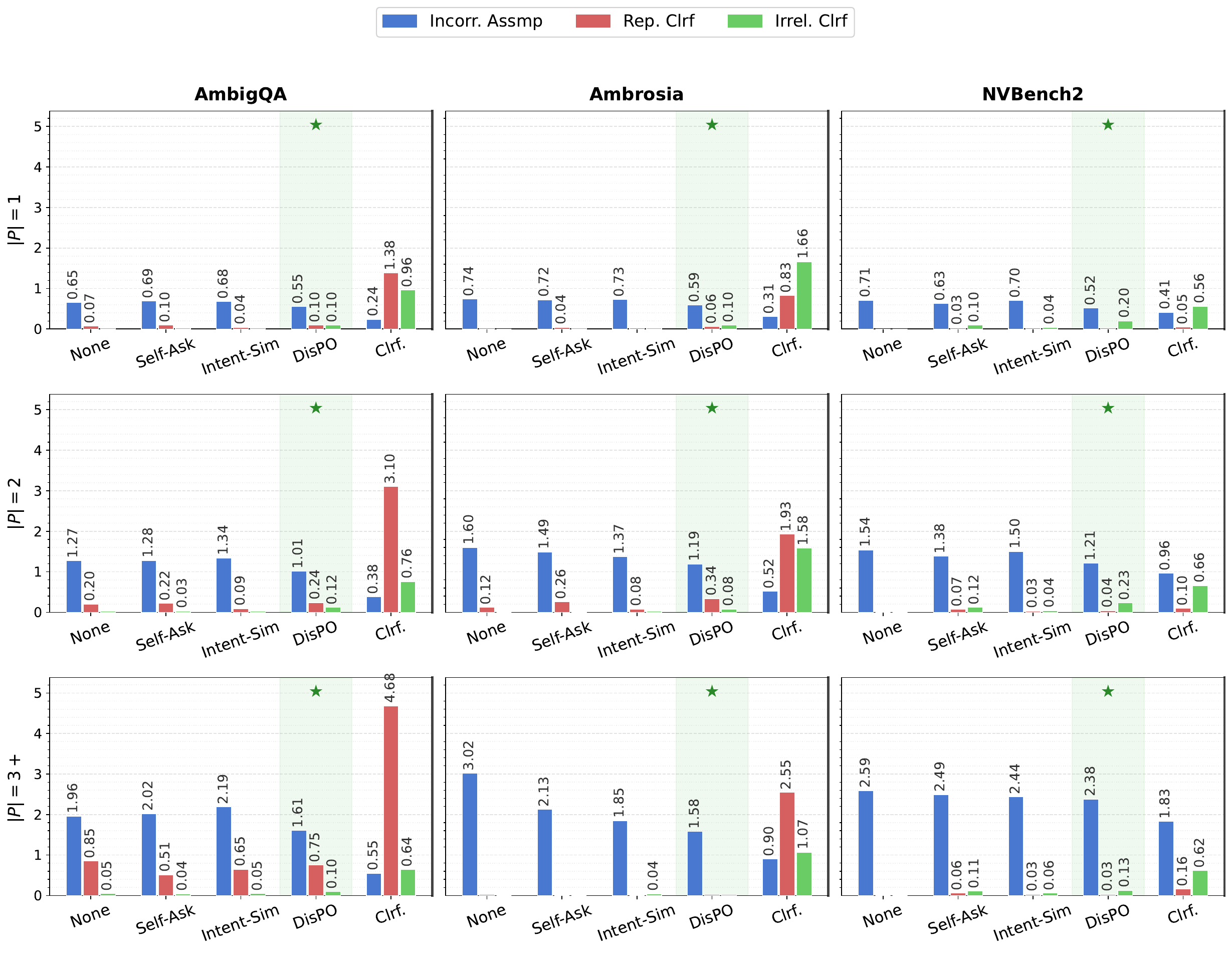}
  \caption{
  \textbf{Variation of dissatisfaction signals as extent of ambiguity $|P|$ increases going vertically down.} 
  Individual bars report average incorrect assumptions (\textcolor{blue}{blue}), repeating clarifications (\textcolor{red}{red}), and irrelevant clarifications (\textcolor{green}{green}). \textsc{None} incurs incorrect assumptions that worsen monotonically with $|P|$, while \textsc{Clrf} observes reduction in IrrelClrf with $|P|$. \textsc{DisPO} achieves a Pareto-optimal profile across all three metrics and ambiguity depth.}
  \label{fig:amb_depth_analysis}
\end{figure*}

Figure~\ref{fig:amb_depth_analysis} shows how dissatisfaction signals get affected across different ambiguity resolution strategies a extent of ambiguity $|P|$ increases.
Incorrect assumption across all strategies increase, suggesting that there is more room for getting an interpretation wrong if the agent is making an assumption.
Same goes for repeating clarification, as the model now has a wider base where its questions resolved earlier resurface in a later turn. 
However the trend is inconsistent as repetition occurs minimally in Ambrosia and nvBench~2.0, that have tasks more goal-oriented than AmbigQA, and therefore require fewer exploratory clarification turns. 
Conversely, irrelevant clarifications decrease slightly as $|P|$ increases, suggesting that clarifying questions are more likely to align with at least one plausible interpretation as the interpretation space expands, leaving fewer questions truly irrelevant.
The DisPO mode achieves a Pareto-optimal profile across all three metrics and all depth levels, consistently maintaining lower incorrect assumptions than solution-driven modes while avoiding the clarification degradation exhibited by Clrf.
Across datasets, repetitive and misdirected clarifications are substantially more common in AmbigQA and Ambrosia than in nvBench~2.0. This is largely because nvBench~2.0 involves structured chart-generation tasks with ambiguity constrained on tables, whereas AmbigQA contains open-domain questions with broader interpretation spaces and ever-changing world-knowledge dependencies.

\subsection{Simulated User Benefits (RQ4)}
\label{sec:user-proxy-ablation}


We compare our simulator against the single-LLM design standard in multi-turn evaluation \cite{collabllm2025}. Because prior simulators were built for different tasks and settings (Section~\ref{sec:related_works}), reusing their prompts directly would confound the comparison with domain mismatch. Instead, we unify the instructions they share into a single generic baseline, keeping those that apply to ambiguity resolution (Appendix~\ref{appx:prompts_used}). This keeps the comparison on methodology, an unconstrained LLM user versus a preference-grounded one, rather than on reproducing prior experimental setups.

\begin{wraptable}{r}{0.52\columnwidth}
\vspace{-10pt}
\centering
\scriptsize
\setlength{\tabcolsep}{3pt}
\begin{tabular}{l cc cc}
\toprule
 & \multicolumn{2}{c}{GPT-5} & \multicolumn{2}{c}{Phi-4} \\
\cmidrule(lr){2-3} \cmidrule(lr){4-5}
Metric & Base. & Ours & Base. & Ours \\
\midrule
Acc (\%)
& \makecell{63.21\\$\pm$6.89}
& \makecell{61.71\\$\pm$0.93}
& \makecell{45.39\\$\pm$14.48}
& \makecell{62.16\\$\pm$0.44} \\

$|$Conv$|$
& \makecell{1.98\\$\pm$0.02}
& \makecell{2.23\\$\pm$0.31}
& \makecell{2.56\\$\pm$0.05}
& \makecell{3.02\\$\pm$0.02} \\
\midrule
Pref. Viol. (\%) & 5.26 & 1.01 & 7.05 & 4.87 \\
Hallucin. (\%)   & 0.00 & 0.00 & 11.00 & 0.00 \\
Overdisc. (\%)   & 3.06 & 1.68 & 2.67 & 0.33 \\
\bottomrule
\end{tabular}
\caption{User simulator comparison.}
\label{tab:user-proxy-ablation}
\vspace{-8pt}
\end{wraptable}


\paragraph{Experimental Setup.}
We use the same sample set employed for DisPO and run simulations with both our simulator and the baseline under the \textit{None} and \textit{Clrf} strategies across all benchmarks using GPT-5.4, resulting in a total of 600 simulations per simulator. 
We further extend the comparison by evaluating both Phi-4 and GPT-5 as the underlying LLMs driving the respective user simulators.
We use LLM-as-judge \cite{collabllm2025} to compute Preference Violation, Hallucination and Over-disclosure rate (all prompts in Appendix~\ref{appx:prompts_used}).





\paragraph{Results.}
Table~\ref{tab:user-proxy-ablation} highlights qualitative and statistical benefits offered by our Preference-Grounded User Simulator.
Accuracy for the baseline user simulator varies significantly with the choice of underlying LLM, suggesting that the evaluated GPT-5.4 agent degrades when paired with less capable simulator models. In contrast, our simulator remains robust across changes in the underlying model, indicating broader applicability without dependence on access to highly capable proprietary systems.
The baseline simulator also exhibits substantially higher standard deviation in accuracy, indicating inconsistent responses across iterations that are difficult to reproduce reliably for scientific evaluation. Additionally, conversations with our simulator tend to be longer due to lower rates of over-disclosure. The baseline, by contrast, often leaks explicit hints or partially solves the task on behalf of the evaluated model, enabling faster convergence to an intent-aligned response with fewer turns. This behavior is further reflected in its higher preference violation rate, where conversations terminate early despite the final solution not aligning with the user’s intended interpretation.


\section{Related Works}
\label{sec:related_works}

\paragraph{Ambiguity Resolution.}
Ambiguity has been studied across many domains \cite{ivanova-etal-2025-ambik}. Prior work commonly treats clarification as an effective resolution strategy \cite{Yu2019InteractiveCB, kuhn2023clamselectiveclarificationambiguous}, overlooking the prolonged, frustrating conversations it can cause. Existing evaluations largely measure how well a model enumerates or covers all interpretations, using beam search or metrics such as F1 \cite{zhang2023clarifynecessaryresolvingambiguity, bhaskar-etal-2023-benchmarking, min2020ambigqaansweringambiguousopendomain}, which limits their applicability when ambiguity is resolved through user interaction \cite{collabllm2025, zhang2025modelingfutureconversationturns}. We extend this line of work by linking open-ended user dissatisfaction signals \cite{inproceedings} to resolution inefficiency, bridging static evaluation and real-world conversational utility.

\paragraph{User Simulator.}
Prior simulators typically rely on a single LLM call, prompted to emulate a human's answering style \cite{collabllm2025}, role-play personas or domain characteristics \cite{wu-etal-2025-aligning, zhao2025personalensbenchmarkpersonalizationevaluation}, or operate on constrained knowledge \cite{shea2026sagetopdownbottomupknowledgegrounded}. They are often further steered toward the final solution with hints \cite{vijayvargiya2025interactiveagentsovercomeambiguity} or intermediate reasoning steps \cite{dutta2025condabenchinteractiveevaluationlanguage}. In contrast, our simulator restricts its responses to user preferences critical to disambiguate the query, improving reproducibility and extensibility across models.

\section{Conclusion}
We introduce a preference-grounded user simulator for controlled and reproducible evaluation of ambiguity resolution in multi-turn interactions, reducing hallucination and over-disclosure common in unconstrained simulators. We further define dissatisfaction signals: incorrect assumptions, repeating clarifications, and irrelevant clarifications, to capture observable inefficiencies during ambiguity resolution. Across question answering, semantic parsing, and text-to-viz. tasks, we reveal clear trade-offs between solution-driven and clarification-driven strategies. DisPO, optimized on dissatisfaction signals, achieves a Pareto-optimal balance, demonstrating what models should strive for in their attempt to ask precisely and assume wisely.

\section{Limitations}
Our simulated user framework is inherently dependent on the quality and completeness of the disambiguation preferences generated. 
Ambiguity in benchmarks such as AmbigQA, Ambrosia, and NvBench~2.0 is well-bounded through high-quality disambiguated queries, enabling the construction of reliable preference sets.
However, extending this process to naturally occurring ambiguous queries remains challenging, as they can admit a large or unbounded number of interpretations, which would trivially reduce IrrelClrf to zero. To address this, we bound the interpretation space by prioritizing critical and commonly intended interpretations over peripheral ones, though this introduces a degree of subjectivity in determining what constitutes a valid or popular interpretation.
Nevertheless unbounded ambiguity space can be modeled in this framework and can be run against a specific user preference set of interest.

Additionally, we do not consider single-turn approaches in which the model generates multiple interpretations alongside their corresponding answers simultaneously in the same turn. Our primary objective is to evaluate conversational efficiency, so we treat such multi-solution outputs as equivalent to separate dialogue turns, each resolving a distinct interpretation. In this setting, the inefficiency of multi-answer generation appears as extra token and computation cost within a single turn rather than as extra turns, and each generated answer can be read as an individual model assumption, which reduces the evaluation to the None setting.

\section{Ethical Considerations}

We use AmbigQA \cite{min2020ambigqaansweringambiguousopendomain}, Ambrosia \cite{saparina2024ambrosiabenchmarkparsingambiguous} and nvBench~2.0 \cite{luo2026nvbench20resolvingambiguity} for evaluation which are open-source under the license CC-BY-4.0\footnote{\url{https://creativecommons.org/licenses/by/4.0/}}.
The user simulator introduced in this work operates under a very strong assumption that its vocabulary is restricted to the disambiguation preferences set for a given interpretation.
While this enables controlled and reproducible evaluation of ambiguity resolution in multi-turn interactions, it does not fully reflect the variability and unpredictability of real user behavior.
Our notion of a user simulator is intended only as an abstraction for studying human-LLM interactions under ambiguity, rather than as a faithful attempt to replicate real users.
Moreover, the dissatisfaction signals derived from this setup may not fully capture the degree of dissatisfaction a real user may experience during such interactions. We instead treat them as immediate observable proxies, derived from our simulator design, for measuring the effectiveness of ambiguity resolution in interactive settings.
This was further approved by the ethics review board by the anonymous IT company. 
None of the interpretations in this study had implication of bias towards specific race, gender, ethnicity, or other protected attributes.
ChatGPT was used solely for grammar polishing and improving writing clarity during the preparation of this paper. It was not involved in the formulation of hypotheses, experimental design, analysis, or the scientific discoveries and conclusions presented in this work.

\bibliographystyle{named}
\bibliography{citation}


\clearpage

\appendix

\section{Dissatisfaction Signals Formulation}
In this section, we discuss in detail how the formulas projected in Section~\ref{sec:dissat_formulas} are justified.

\paragraph{Incorrect Assumption.}
We compute the sum across all turns in a conversation for scenarios when the user responded with a subset of $P$, when no clarification question was asked by the model. This implies that the user, if not ending conversation, is unhappy with misaligned interpretations and applies penalty for the ambiguity dimensions it replies with.

\paragraph{Repeating Clarification.} 
is measured by counting the occurrences of those questions at a given turn that have either been asked or clarified by the user in their response in prior turns. The search space spans both relevant $P$ and irrelevant preferences $\overline{P}$.

\paragraph{Irrelevant Clarification.}
is measured by the number of occurrences of those questions that do not resolve any ambiguity resolved by any preference in $P$. 
In the absence of scaling, a model would be penalized more towards asking questions, thereby creating bias towards making assumptions. It is for this reason we allow some breathing space for models initially to ask questions and increase the penalty as information gets revealed over the conversation.
We normalize each question \(Q(t,p\) for a given irrelevant preference $p$ at turn $t$ with $\kappa^{|P|-r(t)}$, where $\kappa$ is a hyperparameter loosely approximating the number of interpretations for a given ambiguity dimension, ranging somewhere between $2-3$. For example, "How many goals did Ronaldo score?" can either be interpreted as \textit{Christiano Ronaldo} from Portugal or \textit{Ronaldo Nazario} from Brazil.
We use an exponential scaling since every time an ambiguity is resolved, the number os answers in the solution space contracts by $\kappa$.
We further translate $\kappa$ constant and represent it as number of interpretations $\mathcal{I}(q)=\kappa^{|P|}$ for a given query. Hence the normalization factor $\kappa^{|P|-r(t)} \rightarrow \mathcal{I}(q)^{1-\frac{r(t)}{|P|}}$.
Different queries with different $\mathcal{I}(q)$, would have different appetite for accepting irrelevant clarifications.

\section{Qualitative Analysis}
Qualitative analysis reveals recurring patterns across all dissatisfaction signals.

\paragraph{Incorrect Assumptions.}
\begin{itemize}
    \item[\textbf{W-1}] \textbf{Default entity trap:} The model defaults to globally dominant interpretations (e.g., US over UK, film over radio), overriding explicitly intended non-default meanings.
    
    \item[\textbf{W-2}] \textbf{Ordinal/positional default:} Ordinal constraints such as ``second win'' or ``episode 11'' are ignored in favor of first, latest, or most common interpretations.
    
    \item[\textbf{W-3}] \textbf{Granularity mismatch:} The model returns answers at an incorrect granularity, such as returning a team instead of a league or a captain instead of a vice-captain.
    
    \item[\textbf{W-4}] \textbf{Temporal cutoff violations:} Temporal constraints such as ``as of 2017'' are ignored or only partially enforced.
    
    \item[\textbf{W-5}] \textbf{Axis/channel guessing:} In NvBench2.0, models guess chart axis mappings instead of requesting clarification when mappings are ambiguous.
    
    \item[\textbf{W-6}] \textbf{Sample-as-dataset confusion:} Models incorrectly treat example rows as the complete dataset and reject valid chart requests because queried values are absent from the examples.
\end{itemize}

\paragraph{Repeating Clarifications.}
\begin{itemize}
    \item[\textbf{R-1}] \textbf{Over-decomposition cascade:} After a user narrows an ambiguity dimension (e.g., interpreting ``why'' as capability rather than motivation), the model repeatedly sub-divides the same dimension into finer variants, leading to long clarification loops.
    
    \item[\textbf{R-2}] \textbf{Confirmation-seeking:} The model correctly infers the user's intent but still asks redundant confirmation questions such as ``is this what you mean?'' instead of proceeding.
    
    \item[\textbf{R-3}] \textbf{Multi-spec partial resolution:} In queries containing multiple specifications, the model resolves only one ambiguity at a time while repeatedly revisiting previously resolved dimensions.
    
    \item[\textbf{R-4}] \textbf{Lexical re-opening:} Previously resolved ambiguities, such as chart axis assignments, are reopened under different wording or interaction contexts.
\end{itemize}

\paragraph{Irrelevant Clarifications.}
\begin{itemize}
    \item[\textbf{I-1}] \textbf{Aggregation over-asking:} Models ask unnecessary aggregation questions (e.g., ``sum, average, or count?'') even when aggregation intent is already implied.
    
    \item[\textbf{I-2}] \textbf{Unsolicited filter questions:} The model introduces additional filtering options unrelated to the ambiguity specified in the query.
    
    \item[\textbf{I-3}] \textbf{Questioning the chart type:} Instead of resolving missing information, the model questions whether the requested chart type is appropriate.
    
    \item[\textbf{I-4}] \textbf{Format vs. content confusion:} The model asks formatting-related questions when the ambiguity concerns factual content.
    
    \item[\textbf{I-5}] \textbf{Re-opening specified entities:} Explicitly provided entities are unnecessarily questioned instead of being treated as fixed constraints.
\end{itemize}

\section{Statistical Significance Tests}
\label{app:significance}

We rerun GPT-5.4 (chat) with DisPO for five independent iterations on AmbigQA, Ambrosia, and nvBench2.0, and report the standard deviation and variance of the resulting metrics, averaged across the three benchmarks, in Table~\ref{tab:statistical_significance_test}. Accuracy and CIR are reported in \%.

\begin{table}[h]
\small
\centering
\begin{tabular}{lrr}
\toprule
\textbf{Metric} & \textbf{Std Dev} & \textbf{Variance} \\
\midrule
Acc(\%)    & 0.92   & 0.85     \\
$|$Conv$|$ & 0.0515 & 0.003375 \\
IncAssmp   & 0.0540 & 0.003579 \\
RepClrf    & 0.0208 & 0.000561 \\
IrrelClrf  & 0.0478 & 0.003311 \\
CIR(\%)    & 2.76   & 9.77     \\
\bottomrule
\end{tabular}
\caption{Standard deviation and variance of GPT-5.4 with DisPO over five independent runs, averaged across AmbigQA, Ambrosia, and nvBench2.0.}
\label{tab:statistical_significance_test}
\end{table}

The low variation across runs indicates that the trends in Table~\ref{tab:main_table_models} and Table~\ref{tab:main_table_strategy} are stable.


\section{Prompts Used}
\label{appx:prompts_used}

We list down all the system prompts used across all LLM operated systems in this paper.


\begin{figure*}
\begin{tcolorbox}[promptbox, title= AmbigQA System Prompt]
\small
You are a helpful AI assistant. When providing an answer to the user's question, you think out loud and reason step by step. You clearly mention any assumptions that were made while answering the question.

\medskip
\begin{tcolorbox}[insertbox=gray, title=$\hookrightarrow$ Nature Insert: \textbf{(C)} or \textbf{(G)} — see above]
\centering\textit{Swapped per evaluation mode}
\end{tcolorbox}

\medskip
Your final answer, which should be a concise one or two words (not a full sentence) should be represented in the following format:
\begin{verbatim}
```json
{
  "answer": "your final answer here"
}
```
\end{verbatim}

This should follow after you have stated your reasoning and assumptions (if any) for the final answer provided. For example,

\textit{query:} Who is the current chairman of african union commission?\\
\textit{response:} As of my latest information (through Sept 2025), the fourth and the current Chairperson of the African Union Commission is Moussa Faki Mahamat from Chad.
\begin{verbatim}
```
{
  "answer": "Moussa Faki Mahamat"
}
```
\end{verbatim}

If asking for clarification, simply respond in plain text without using the \texttt{json} tags. For example,

\textit{query:} Who is the current chairman of african union commission?\\
\textit{response:} Are you asking about the current chairman as of 2025 or any other year?

You should use \texttt{json} tags only when you are providing a final answer to the user's query. If you are asking for clarifications, do not use the \texttt{json} tags and just ask the question in plain text.
\end{tcolorbox}
\end{figure*}


\begin{figure*}
\begin{tcolorbox}[promptbox, title= Ambrosia System Prompt]
\small
The task is to write SQL queries based on the provided questions in English. Do not include any explanations, and do not select extra columns beyond those requested in the question.

\medskip
\begin{tcolorbox}[insertbox=gray, title=$\hookrightarrow$ Nature Insert: \textbf{(C)} or \textbf{(G)} — see above]
\centering\textit{Swapped per evaluation mode}
\end{tcolorbox}

\medskip
If you are providing a solution, enclose the SQL query inside \verb|```sql```| tags. If you are providing feedback or asking for clarification, respond in plain text without any code formatting. Presence of \verb|```sql```| tags will imply that you are providing a solution.
\end{tcolorbox}
\end{figure*}


\begin{figure*}
\begin{tcolorbox}[promptbox, title=nvBench2.0 System Prompt]
\small
You are an expert data visualization assistant. Given a natural language visualization request and a data table schema with example values, generate a Vega-Lite chart specification. Only use column names that exist in the provided schema. Do not include any explanations or extra fields beyond what is needed to represent the chart.

\medskip
\begin{tcolorbox}[insertbox=gray, title=$\hookrightarrow$ Nature Insert: \textbf{(C)} or \textbf{(G)} — see below]
\centering\textit{Swapped per evaluation mode}
\end{tcolorbox}

\medskip
If you are providing a solution, enclose your answer in \verb|```json```| tags using the following output format:
\verb|```json|\\
\texttt{\{chart\_spec\}}\\
\verb|```|

Each Vega chart must only include the following three keys (\texttt{mark}, \texttt{encoding}, \texttt{transform}):

\textbf{mark} (chart type)
\begin{itemize}
  \item Possible chart types: \texttt{bar, line, arc, point, rect, boxplot}. Note that \texttt{arc} is identical to pie chart, and \texttt{rect} is identical to heatmap.
  \item If the NL query mentions one exact chart type, select that single chart type.
  \item Else if the NL query does not mention chart type but indicates a data analysis task in (trend; distribution; correlation), infer the chart type by (task:chart) = (trend:[bar,line]; distribution:[bar,arc,line,boxplot]; correlation:[point,heatmap]).
  \item Otherwise, when no hints for chart type are mentioned in the NL query, all chart types are possible.
\end{itemize}

\textbf{encoding} (with only \texttt{field}, \texttt{aggregate}, \texttt{sort}, \texttt{bin}, and \texttt{timeUnit} within the channels)
\begin{itemize}
  \item Obligatory/optional channels per chart type (channels with * are obligatory): bar:[x*,y*,color]; line:[x*,y*,color]; arc:[color*,theta*]; point:[x*,y*,color,size]; rect:[x*,y*,color*]; boxplot:[x*,y*].
  \item \texttt{aggregate}: Three types: \texttt{sum}, \texttt{mean}, \texttt{count}. For \texttt{count}, it is a special computed column that must not include a \texttt{field}.
  \item \texttt{sort}: If sort is explicitly requested, use channels like \texttt{x}, \texttt{-x}, \texttt{y}, \texttt{-y}, \texttt{theta}, or \texttt{-theta}.
  \item \texttt{bin}: Used for quantitative columns, e.g., \texttt{\{"bin": \{"maxbins": 10\}\}}. For temporal columns, use \texttt{timeUnit}, e.g., \texttt{\{"timeUnit": "year"\}}.
\end{itemize}

\textbf{transform}
\begin{itemize}
  \item Include only \texttt{filter} if explicitly mentioned in the NL query; otherwise omit \texttt{transform}.
\end{itemize}

\medskip
If you need clarification or are providing feedback, respond in plain text without any code formatting. The presence of \verb|```json```| tags indicates you are submitting a solution.
\end{tcolorbox}
\end{figure*}


\begin{figure*}
\begin{tcolorbox}[insertbox=blue, title=Nature Insert \textbf{(C)}: \texttt{clarification} mode]
\small
\textbf{Your Nature:}
The queries that you will face would be ambiguous. Always clarify all the ambiguities by asking relevant clarification questions to the user before providing an answer. Do not make any assumptions.
\end{tcolorbox}
\label{fig:clarification_prompt}
\end{figure*}

\begin{figure*}
\begin{tcolorbox}[insertbox=teal, title=Nature Insert \textbf{(G)}: \texttt{DisPO} mode]
\small
\textbf{Your Nature:}
The queries you will face are ambiguous. Ask clarifying questions to resolve genuine ambiguities before answering. Follow these rules strictly:
\begin{enumerate}
  \item \textbf{Ask only about genuine ambiguities.} Do not ask about anything already specified or implied by the query, or that has a clear conventional default.
  \item \textbf{Accept every clarification as final.} Once the user answers, act on it immediately --- never rephrase the same question or ask ``just to confirm.''
  \item \textbf{Do not sub-divide a resolved dimension.} If the user narrows an ambiguity to a category, use your best judgment within that category rather than decomposing it further.
  \item \textbf{Do not default to the most prominent interpretation.} If a query is ambiguous between a well-known reading and a less common one, ask --- do not silently assume the most familiar one.
  \item \textbf{Apply every constraint in the query.} Never silently drop a qualifying constraint (temporal, numerical, geographic, categorical, etc.).
  \item \textbf{Treat provided examples as illustrative, not exhaustive.} Do not refuse or flag errors because a query value is absent from example data --- trust the query and proceed.
  \item \textbf{Infer well-established conventions.} Apply standard, widely accepted approaches without asking; clarify only when the choice meaningfully changes the outcome.
  \item \textbf{Batch all ambiguities into one turn.} Ask about all unresolved dimensions at once, then produce your response as soon as all are answered.
\end{enumerate}
\end{tcolorbox}
\end{figure*}


\begin{figure*}
\label{fig:baseline_user_proxy}
\begin{tcolorbox}[promptbox, title=Baseline User Proxy System Prompt]
\small
You are simulating a user in a conversation with an AI assistant.

\medskip
\textbf{Context}\\
The user has an ambiguous query and a set of preferences that resolve the ambiguity. The AI assistant is trying to help the user.

\medskip
\textbf{Your Task}\\
Given the conversation history, you must produce the next user reply. Follow these rules strictly:
\begin{enumerate}
  \item If the assistant's latest message contains a final answer (e.g.\ a SQL query, a code block, a direct answer), inspect whether it matches all of the user's preferences.
  \begin{itemize}
    \item If all preferences are satisfied, reply \textbf{only} with: \textit{``Thanks! I am satisfied with your solution''}
    \item If any preference is violated, point out the violated preference(s) and ask the assistant to try again.
  \end{itemize}
  \item If the assistant's latest message is asking a clarification question, answer it using the relevant preference(s) that address the ambiguity being asked about.
  \begin{itemize}
    \item If none of the preferences address the clarification question, reply: \textit{``I have no preferences. Please proceed with your best judgement''}
  \end{itemize}
\end{enumerate}

\medskip
\textbf{Input}
\begin{itemize}
  \item \texttt{query}: The original ambiguous question from the user.
  \item \texttt{preferences}: The user's preferences for resolving the ambiguity, as a list.
  \item \texttt{conversation}: The full conversation history up to and including the assistant's latest message.
\end{itemize}

\medskip
\textbf{Behavioral Guidelines}
\begin{itemize}
  \item Stay in character as a human user. You are NOT an AI. Do not respond like an assistant.
  \item Goal-oriented: keep the conversation focused on resolving the original query. Avoid small talk or digressions.
  \item Keep your responses concise and to the point.
\end{itemize}

\medskip
\textbf{Output}\\
Produce only the user's next reply as a short plain-text message. Do not include any explanation or meta-commentary.
\end{tcolorbox}

\end{figure*}


\begin{figure*}
\begin{tcolorbox}[title=Hallucination Judge System Prompt, fonttitle=\bfseries]
You are an expert evaluator assessing whether a simulated USER in a multi-turn conversation has hallucinated.

\textbf{Definition.} A hallucination occurs when the simulated user:
\begin{enumerate}
    \item Claims an action was performed (by themselves or the agent) when it was not.
    \item Claims an action was \emph{not} performed when it actually was.
    \item Makes a request or complaint that has no relationship to any of the stated user preferences.
\end{enumerate}

\textbf{Input}
\begin{itemize}
    \item \texttt{query}: The original ambiguous question.
    \item \texttt{amb\_specs}: The list of user preferences that resolve the ambiguity.
    \item \texttt{conversation}: The full conversation showing both user and assistant turns.
\end{itemize}

\textbf{Task.} Review only the USER turns and determine whether any of the three hallucination types occurred.

\textbf{Output} --- strict JSON:
\begin{verbatim}
{ "hallucinated": true | false, "reasoning": "<brief explanation>" }
\end{verbatim}
\end{tcolorbox}
\end{figure*}


\begin{figure*}
\begin{tcolorbox}[title=Over-Disclosure Judge System Prompt, fonttitle=\bfseries]
You are an expert evaluator assessing whether a simulated USER in a multi-turn conversation over-disclosed information that should only be discovered by the AI agent.

\textbf{Definition.} Over-disclosure occurs when the simulated user:
\begin{enumerate}
    \item Attempts to solve the question themselves (e.g.\ computes or constructs the answer).
    \item Leaks the correct answer or a critical hint that makes the solution too obvious (e.g.\ reveals the specific column name, value, or formula the agent should discover).
    \item Gives away the intended resolution in a way that circumvents the agent's reasoning process.
\end{enumerate}

\textbf{Note:} It is acceptable for the user to answer clarification questions using the provided preferences. That is \emph{not} over-disclosure.

\textbf{Input}
\begin{itemize}
    \item \texttt{query}: The original ambiguous question.
    \item \texttt{amb\_specs}: The list of user preferences that resolve the ambiguity.
    \item \texttt{conversation}: The full conversation showing both user and assistant turns.
\end{itemize}

\textbf{Task.} Review only the USER turns and determine whether over-disclosure occurred.

\textbf{Output} --- strict JSON:
\begin{verbatim}
{ "overdisclosed": true | false, "reasoning": "<brief explanation>" }
\end{verbatim}
\end{tcolorbox}
\end{figure*}


\section{Within-Query Sensitivity Analysis}
\label{app:sensitivity}

To test how sensitive our metrics are to the choice of user preference set $P$, we measure within-query variation across interpretations for GPT-5.4. For each query, we compute the standard deviation of each metric across its interpretations (i.e., across different assignments of $P$), and report the mean of these per-query standard deviations in Table~\ref{tab:sensitivity}. Higher values indicate greater variation across interpretations. Accuracy and CIR are binary (0/1) per conversation, IncAssmp, RepClrf and IrrelClrf are raw counts, and $|\text{Conv}|$ is the number of turns.

\begin{table*}[t]
\centering
\small
\setlength{\tabcolsep}{4pt}
\begin{tabular}{llcccccc}
\toprule
\textbf{Data} & \textbf{Strategy} & \textbf{Acc} & \textbf{IncAssmp} & \textbf{RepClrf} & \textbf{IrrelClrf} & \textbf{CIR} & $|\textbf{Conv}|$ \\
\midrule
\multirow{5}{*}{AmbigQA}
 & None       & 0.30 & 0.52 & 0.17 & 0.02 & 0.43 & 0.62 \\
 & Self-Ask   & 0.29 & 0.53 & 0.19 & 0.02 & 0.39 & 0.62 \\
 & Intent-Sim & 0.31 & 0.54 & 0.12 & 0.03 & 0.42 & 0.58 \\
 & DisPO      & 0.30 & 0.51 & 0.23 & 0.10 & 0.40 & 0.67 \\
 & Clrf       & 0.32 & 0.28 & 1.78 & 0.65 & 0.28 & 1.68 \\
\midrule
\multirow{5}{*}{Ambrosia}
 & None       & 0.24 & 0.77 & 0.03 & 0.00 & 0.47 & 0.71 \\
 & Self-Ask   & 0.22 & 0.78 & 0.07 & 0.02 & 0.48 & 0.77 \\
 & Intent-Sim & 0.22 & 0.74 & 0.04 & 0.02 & 0.48 & 0.73 \\
 & DisPO      & 0.22 & 0.69 & 0.09 & 0.10 & 0.43 & 0.71 \\
 & Clrf       & 0.23 & 0.40 & 1.15 & 0.76 & 0.14 & 1.16 \\
\midrule
\multirow{5}{*}{nvBench2.0}
 & None       & 0.09 & 0.53 & 0.00 & 0.00 & 0.33 & 0.37 \\
 & Self-Ask   & 0.09 & 0.53 & 0.02 & 0.06 & 0.33 & 0.39 \\
 & Intent-Sim & 0.09 & 0.55 & 0.04 & 0.06 & 0.34 & 0.46 \\
 & DisPO      & 0.10 & 0.50 & 0.02 & 0.15 & 0.30 & 0.37 \\
 & Clrf       & 0.10 & 0.43 & 0.09 & 0.35 & 0.25 & 0.40 \\
\bottomrule
\end{tabular}
\caption{Within-query sensitivity of GPT-5.4 across ambiguity-resolution strategies. Values are the mean per-query standard deviation across interpretations; higher values indicate more variation across user preference sets $P$.}
\label{tab:sensitivity}
\end{table*}

Accuracy is most sensitive to the interpretation on AmbigQA (0.30) and least on nvBench2.0 (0.09). CIR varies substantially across interpretations on all datasets (roughly 0.3--0.5 for non-clarifying strategies), indicating that interaction cleanliness depends on which interpretation the user holds. Clarification increases the variability of repeated and irrelevant clarifications and of conversation length, while reducing the variability of IncAssmp and CIR. IncAssmp is most volatile for strategies that do not ask (None, Self-Ask, Intent-Sim), suggesting that a single assumption may fit one interpretation but not others.



\end{document}